\documentclass{article}

\usepackage{tabulary,graphicx,times,caption,fancyhdr,amsfonts,amssymb,amsbsy,latexsym,amsmath}
\usepackage[utf8]{inputenc}
\usepackage{url,multirow,morefloats,floatflt,cancel,tfrupee,textcomp,colortbl,xcolor,pifont}
\usepackage[nointegrals]{wasysym}

\usepackage{amsthm}
\newtheorem{Definition}{Definition}[section]
\newtheorem{Remark}{Remark}[section]
\newtheorem{Assumption}{Assumption}[section]
\newtheorem{Example}{Example}[section]
\newtheorem{Proposition}{Proposition}[section]
\newtheorem{Lemma}{Lemma}[section]

\newtheorem{Theorem}{Theorem}[section]

\usepackage{multicol}

\makeatletter

\usepackage{ifxetex}
\ifxetex\else
  \usepackage{dblfloatfix}
\fi

\@ifundefined{subparagraph}{
\def\subparagraph{\@startsection{paragraph}{5}{2\parindent}{0ex plus 0.1ex minus 0.1ex}%
{0ex}{\normalfont\small\itshape}}%
}{}

\def\URL#1#2{\@ifundefined{href}{#2}{\href{#1}{#2}}}

\def\UrlOrds{\do\*\do\-\do\~\do\'\do\"\do\-}%
\g@addto@macro{\UrlBreaks}{\UrlOrds}

\makeatother

\usepackage[paperheight=11in,paperwidth=8.3in,margin=2.5cm,headsep=.7cm,top=2.5cm]{geometry}
\usepackage[T1]{fontenc}

\renewenvironment{abstract}
	{\trivlist\item[]\leftskip0pt\par\vskip4pt\noindent
  	\textbf{\abstractname}\mbox{\null}\\}
	{\par\noindent\endtrivlist}

\def\keywords#1{\par\medskip\par\noindent\textbf{Keywords}: #1\par}

 \date{} \emergencystretch 8pt

\makeatletter
\def\author#1{\gdef\@author{\hskip-\tabcolsep%
	\parbox{\textwidth}{\raggedright\bfseries#1\\[1pc]}}}
\def\address[#1]#2{\g@addto@macro\@author{\\\hskip-\tabcolsep\parbox{\textwidth}{\raggedright%
	\normalsize\normalfont\textsuperscript{#1}#2}}}
\let\addresslink\textsuperscript
\def\correspondence#1{\g@addto@macro\@author{\\\hskip-\tabcolsep\parbox{\textwidth}{\raggedright%
	\vspace*{10pt}\normalsize\normalfont~\\#1~\\[12pt]}}}
\def\email#1{\g@addto@macro\@author{\\\hskip-\tabcolsep\parbox{\textwidth}{\raggedright%
	\normalsize\normalfont  #1}}}

\def\title#1{\gdef\@title{\vspace*{-30pt}%
	\raggedright\textbf{\@journaltitle}~\\%
  \raggedright\bfseries\ifx\@articleType\@empty\vspace*{20pt}\else%
  \vspace*{20pt}\@articleType\vspace*{20pt}\\\fi#1}}
\let\@journaltitle\@empty \def\journaltitle#1{\gdef\@journaltitle{{\normalfont\itshape#1}}}
\let\@articleType\@empty \def\articletype#1{\gdef\@articleType{{\normalfont\itshape#1}}}

\let\@runningHead\@empty \def\RunningHead#1{\gdef\@runningHead{{\normalfont #1}}}

\usepackage{fancyhdr}
\fancypagestyle{headings}{\fancyhf{}
  \fancyhead[R]{\itshape\@runningHead}
  \fancyfoot[C]{\thepage}}
\fancypagestyle{plain}{%
	\fancyhf{}\rhead{arXiv Preprint}
  \fancyfoot[C]{\thepage}}
\makeatother

\usepackage[%
	numbers,sort&compress%
  ]{natbib}

\usepackage{float,xcolor}

\journaltitle{}
\articletype{Research Article} 

\begin{document}

\title{Game-Theoretical Analysis of Reviewer Rewards in Peer-Review Journal Systems: Analysis and Experimental Evaluation using Deep Reinforcement Learning}

\author{%
		Minhyeok Lee\addresslink{1}
    }
		
\address[1]{School of Electrical and Electronics Engineering, Chung-Ang University, Seoul 06974, Korea}


\email{mlee@cau.ac.kr}%

\RunningHead{Submitted to Journal of Mathematics}

\maketitle


\begin{abstract}
In this paper, we navigate the intricate domain of reviewer rewards in open-access academic publishing, leveraging the precision of mathematics and the strategic acumen of game theory. We conceptualize the prevailing voucher-based reviewer reward system as a two-player game, subsequently identifying potential shortcomings that may incline reviewers towards binary decisions. To address this issue, we propose and mathematically formalize an alternative reward system with the objective of mitigating this bias and promoting more comprehensive reviews. We engage in a detailed investigation of the properties and outcomes of both systems, employing rigorous game-theoretical analysis and deep reinforcement learning simulations. Our results underscore a noteworthy divergence between the two systems, with our proposed system demonstrating a more balanced decision distribution and enhanced stability. This research not only augments the mathematical understanding of reviewer reward systems, but it also provides valuable insights for the formulation of policies within journal review system. Our contribution to the mathematical community lies in providing a game-theoretical perspective to a real-world problem and in the application of deep reinforcement learning to simulate and understand this complex system.

\end{abstract}

\section{Introduction}
\label{sec:intro}

The world of academia has witnessed a profound shift with the rise and growth of open-access journals, enabling scientific research to be shared publicly and widely at a scale never seen before \cite{piwowar2018state, martin2018evidence, willinsky2018scholarly}. Nevertheless, the critical need for thorough and impactful peer review continues to be a central challenge in this setting. To inspire reviewers, some open-access journals have introduced a reward system in the form of fee waivers for publishing, offered as vouchers \cite{MDPI}. This paper delves into a thorough mathematical and game-theoretical examination of this voucher-based reviewer reward system.

Under the current system, a tendency exists for reviewers to gravitate towards black-and-white outcomes, that is, either accepting or rejecting a manuscript, instead of proposing amendments or enhancements. This is largely because rewards are typically granted more quickly when a paper is either wholly accepted or rejected, as opposed to when revisions are suggested. Although this method can speed up the review process, it risks compromising the quality and depth of peer reviews, thus promoting an all-or-nothing decision-making approach over a more refined process that includes revisions. This paper proposes and examines a new reward system designed to balance these incentives and promote more thorough reviews.

Our study begins by framing the peer review process as a strategic interaction, or game, between two reviewers. Using the principles of game theory, we carefully analyze the strategies and behaviors displayed by the reviewers when influenced by both the current and proposed reward systems. In our newly proposed system, the reward structure is adjusted to not only promote agreement between reviewers but also to discourage hasty or binary decisions.

To investigate the features of both reward systems, we use deep reinforcement learning techniques to simulate this game \cite{ladosz2022exploration, le2022deep, panzer2022deep}. Two agents, represented as neural networks \cite{wu2020comprehensive, liu2021algorithms, zhang2021survey}, are trained to mimic the roles of reviewers, aiming to maximize their anticipated gain in the game. The training of these agents is steered by a reward function that reflects the voucher reward system common in open-access journals, and a loss function that promotes decisions in line with the true quality of the paper under review.

This paper offers a mathematical explanation of the current reviewer reward system, interpreted as a game. It outlines a model of the current system, followed by a game-theoretical analysis that includes defining dominant strategies and a thorough discussion of reviewer behavior under this system. Furthermore, we propose a novel reviewer reward system based on game-theoretical methods, followed by analyzing the strategies and behaviors it encourages in reviewers. This includes defining optimal strategies and an in-depth look at how these strategies are shaped by the changes we suggest to the reward system.

Throughout this research, we emphasize the game-theoretical foundations of our work. By breaking down the interaction between reviewers as a strategic game, we can meticulously analyze and contrast the two systems in a mathematical framework. This approach allows us to pinpoint potential weaknesses in the existing system and provide solid mathematical evidence supporting the effectiveness of our proposed system.

This paper intends to contribute significantly to the existing body of mathematical literature, offering a fresh game-theoretic perspective on the reviewer reward system in open-access journals. We trust that our findings will encourage further mathematical exploration into this vital aspect of the academic publishing process. Our work may also inform the policies of open-access journals, equipping them with a sound theoretical foundation for crafting reviewer reward systems that endorse thorough and unbiased reviews.

\section{Preliminaries}
\label{sec:preliminaries}
\subsection{Game Theory}
\label{subsec:gametheory}

Game theory is a mathematical framework designed for understanding the interaction among rational agents \cite{bajari2013game, shi2012game, zhou2019survey}. A game in this context is a situation where a number of agents interact, each trying to optimize their own objectives.

\begin{Definition}
\label{def:game}
A game $\mathcal{G}$ is defined by a triplet $(\mathcal{P}, \mathcal{S}, \mathcal{U})$, where:
\begin{itemize}
\item $\mathcal{P} = \{1,2,...,n\}$ is a set of $n$ players.
\item $\mathcal{S} = S_1 \times S_2 \times \cdots \times S_n$ is the strategy space, with $S_i$ being the strategy set of player $i$.
\item $\mathcal{U} = (U_1, U_2, ..., U_n)$ is the set of utility functions, with $U_i: \mathcal{S} \rightarrow \mathbb{R}$ being the utility function of player $i$.
\end{itemize}
\end{Definition}

In this setting, a strategy is a complete plan of action that defines what a player would do in every situation. A strategy profile $s \in \mathcal{S}$ is a vector that specifies a strategy for every player. Each player $i$'s utility function $U_i$ maps from the set of strategy profiles to the real numbers, representing the player's preferences over the set of possible outcomes. 

\begin{Definition}
\label{def:nash}
A strategy profile $s^* \in \mathcal{S}$ is a Nash equilibrium if, for each player $i \in \mathcal{P}$, we have
\begin{equation}
\label{eq:nash}
U_i(s^*_i,s^*_{-i}) \geq U_i(s_i,s^*_{-i}), \quad \forall s_i \in S_i,
\end{equation}
where $s^*_{-i}$ denotes the strategies of all players except player $i$ in strategy profile $s^*$.
\end{Definition}

The Nash equilibrium, as defined in Definition \ref{def:nash}, is a state of the game where no player can unilaterally deviate from their current strategy without decreasing their utility, given that all other players stick to their strategies. This concept is central in game theory and provides a solution concept for strategic interactions.

\begin{Remark}
\label{rmk:mixedstrategies}
In some games, players may not have pure strategy Nash equilibriums, i.e., equilibriums where players use deterministic strategies. In such cases, players may resort to mixed strategies, where they randomize over their pure strategies. A mixed strategy Nash equilibrium is a profile of mixed strategies from which no player wishes to deviate.
\end{Remark}

\begin{Definition}
\label{def:payoffmatrix}
A payoff matrix $\mathcal{M}$ of a two-player game is a matrix where the $(i, j)$-th entry represents the utility of player 1 when player 1 chooses strategy $i$ and player 2 chooses strategy $j$.
\end{Definition}

The payoff matrix is a straightforward representation of a two-player game where each player has a finite number of strategies \cite{schelling2010game, han2012equilibrium, akccay2011evolution}. This matrix offers an efficient method to examine the utilities for different strategy combinations.

\begin{Example}
\label{ex:payoffmatrix}
Consider a two-player game where each player has two strategies. The payoff matrix for player 1 is
\begin{equation*}
\mathcal{M} = \begin{bmatrix} U_1(s_1, s_1) & U_1(s_1, s_2) \\ U_1(s_2, s_1) & U_1(s_2, s_2) \end{bmatrix}.
\end{equation*}
\end{Example}

\begin{Proposition}
\label{prop:dominantstrategy}
In a two-player game, if there exists a strategy $s_i$ for player $i$ such that $U_i(s_i, s_j) \geq U_i(s_k, s_j)$ for all $s_j \in S_j$ and $s_k \in S_i$, $s_k \neq s_i$, then $s_i$ is a dominant strategy for player $i$.
\end{Proposition}

This proposition defines a dominant strategy, which yields the highest utility for a player, regardless of the strategies chosen by other players.

\begin{Lemma}
\label{lem:dominantstrategyNE}
In a two-player game, if there exists a strategy profile $(s^*_1, s^*_2)$ such that $s^*_1$ is a dominant strategy for player 1 and $s^*_2$ is a dominant strategy for player 2, then $(s^*_1, s^*_2)$ is a Nash equilibrium.
\end{Lemma}

This lemma establishes a connection between the dominant strategy and Nash equilibrium in a two-player game. If both players have dominant strategies, the strategy profile consisting of these dominant strategies is a Nash equilibrium.

\begin{Assumption}
\label{asmp:rationality}
We assume that all players are rational, i.e., each player seeks to maximize their own utility.
\end{Assumption}

This assumption is a basic tenet of game theory and is essential for our subsequent analysis. It posits that players will always act in a way that they believe will yield them the greatest utility.

\subsection{Open-Access Journal Review System}
\label{subsec:openaccess}

Open-access (OA) journals have emerged as a significant development in the scholarly publishing industry \cite{albert2006open, solomon2012study, bjork2012open}. These journals provide unrestricted online access to peer-reviewed scholarly research. In the OA model, rather than charging readers to access the content, the publication costs are usually covered by author fees or institutional funding.

\begin{Definition}
\label{def:openaccess}
Open-access publishing is a model for scholarly journal publishing that makes research articles available to readers at no cost, as opposed to the traditional subscription model in which readers have access to scholarly research by paying a subscription fee.
\end{Definition}

The open-access publishing model has revolutionized the way knowledge is disseminated, allowing for a much wider spread of scholarly research. It is predicated on the philosophy of communal sharing of knowledge, making the fruits of academic endeavor available to all, free of charge. The financial burden is thus transferred from the reader to the author, ensuring that the paywalls, which may hinder the proliferation of knowledge, are eliminated.

An integral part of the academic publishing process, irrespective of the publishing model, is the peer-review process. This process is designed to maintain the quality and integrity of scholarly research. In the OA model, the reviewers play a crucial role in assessing the quality of papers.

\begin{Definition}
\label{def:review}
A peer review is an evaluation of scientific, academic, or professional work by others working in the same field to maintain quality and improve performance. In the context of OA publishing, the peer reviewers assess the quality of the submissions and make a recommendation regarding their publication.
\end{Definition}

Peer review acts as a gatekeeper, ensuring that only rigorous and high-quality research is disseminated \cite{ali2016peer, lee2013bias, jefferson2002effects}. The process involves the evaluation of an academic manuscript by experts in the same field as the paper. This stringent process helps to ensure that the research presented is of the highest standard and adds value to the field. The rigorous nature of this process is a critical factor in upholding the reputation and credibility of scientific research.

There are usually three options for the reviewers when assessing a paper: Accept, Revision, and Reject. An "Accept" decision means that the paper is ready for publication in its current form. A "Revision" decision indicates that the paper has potential but requires some changes before it is ready for publication. A "Reject" decision means that the paper is not suitable for publication in the journal.

\subsection{Assumptions in Review System}
\label{subsec:game_form}
\begin{Definition}
\label{def:reviewdecisions}
Let $\mathcal{D} = \{\text{Accept}, \text{Revision}, \text{Reject}\}$ represent the set of possible decisions a reviewer can make for a paper. A decision $d \in \mathcal{D}$ is a recommendation made by a reviewer regarding the publication of a paper.
\end{Definition}

The tripartite system of Accept, Revision, and Reject decisions provides a comprehensive framework for evaluating academic papers. An acceptance indicates a work of high quality, requiring no further alterations. A recommendation for revision, on the other hand, indicates that while the work holds promise, there are aspects that require refinement and improvement before it meets the standards of the journal. A rejection signifies that the work, in its current form, does not meet the necessary standards of quality and contribution to the field.

\begin{Assumption}
\label{asmp:reviewers}
We assume that each paper is reviewed by two independent reviewers, and each reviewer makes a decision from the set $\mathcal{D}$ independently based on their assessment of the paper's quality.
\end{Assumption}

\begin{Remark}
\label{rmk:thirdreviewer}
In the event that the two reviewers provide differing decisions, a third reviewer may be called upon to break the tie. The final decision is then based on the majority opinion.
\end{Remark}

The introduction of a third reviewer in case of a disagreement between the initial reviewers ensures that the decision-making process is fair and unbiased. This mechanism ensures that a single reviewer cannot unduly influence the outcome, thus promoting the integrity of the review process. This procedure also underscores the importance of consensus in academic publishing.

In some OA journals, as a part of the incentive system, reviewers are rewarded with vouchers upon completion of their review tasks. These vouchers can be used for discounts on the open access publishing fees when the reviewers submit their own papers.

\begin{Definition}
\label{def:voucher}
A reviewer reward voucher is a form of remuneration given to reviewers upon completion of a review task. This voucher can be used as a discount on the open-access publishing fees when the reviewers submit their own papers.
\end{Definition}

The voucher system serves as an incentive for reviewers, providing them with tangible recognition for their work. It is a critical aspect of the open-access model, as it helps to ensure a constant supply of reviewers. This system also encourages the reviewers to submit their own work to the journal, creating a cycle of submission and review that sustains the journal.

\begin{Assumption}
\label{asmp:vouchertime}
We assume that the time at which the voucher is received by a reviewer depends on the decision they make. If the decision is either "Accept" or "Reject", the voucher is received immediately after the decision. If the decision is "Revision", the voucher is received after the revised paper has been reviewed.
\end{Assumption}

\begin{Remark}
\label{rmk:discountedvalue}
Given Assumption \ref{asmp:vouchertime}, the voucher value is effectively discounted over time in the case of a "Revision" decision due to the delay in receipt. Consequently, even though the ostensible amount of the voucher is the same for all three decisions, the present value of the voucher for a "Revision" decision is less than that for an "Accept" or "Reject" decision.
\end{Remark}

The temporal discounting of the voucher in case of a "Revision" decision introduces an interesting dynamic into the reviewer's decision-making process. While the face value of the voucher remains the same, the delay in receiving it effectively reduces its value.

The quality of a paper is an inherent characteristic that can be assessed to varying degrees by different reviewers. We denote the true quality of a paper "p" as $S_p$, and we assume that each reviewer has a certain ability to assess this quality with some bias.

\begin{Definition}
\label{def:paperquality}
The quality score of a paper $p$, denoted as $S_p$, is an inherent attribute of the paper that indicates its suitability for publication. Higher scores represent higher quality.
\end{Definition}

The quality score of a paper, though a subjective measure, provides an essential yardstick for evaluation. It encapsulates the perceived merit of the paper, including aspects such as originality, rigor, and contribution to the field. This score, though subjective, is a critical component in the decision-making process of the reviewers.

\begin{Assumption}
\label{asmp:reviewerbias}
The estimated quality score of a paper $p$ by reviewer $i$, denoted as $\hat{S}^{i}_{p}$, is given by $S_p + b^{i}_{p}$, where $b^{i}_{p}$ is the bias of reviewer $i$.
\end{Assumption}

\begin{Assumption}
\label{asmp:privateinformation}
We assume that the estimated quality score $\hat{S}^{i}_{p}$ of a paper by reviewer $i$ is private information, not shared among the reviewers.
\end{Assumption}

This assumption is critical to our model and reflects the real-world situation in the peer review process. Typically, reviewers do not have access to each other's evaluations until after they have submitted their own. This ensures the independence and integrity of the review process.

The estimated quality score $\hat{S}^{i}_{p}$ encapsulates both the inherent quality of the paper $S_p$ and the personal bias of the reviewer $b^{i}_{p}$. The inherent quality $S_p$ is assumed to be a fixed attribute of the paper, known to all reviewers in an ideal world. However, the bias $b^{i}_{p}$ reflects the individual judgment of the reviewer and can vary widely among different reviewers. This variation introduces uncertainty into the review process, as reviewers may disagree about the value of $b^{i}_{p}$ and thus the quality of the paper $\hat{S}^{i}_{p}$.

Given the lack of common knowledge about $\hat{S}^{i}_{p}$, each reviewer must make their decision based on their individual perception of the paper's quality, taking into consideration the potential decisions of the other reviewers. This forms the basis of the strategic interaction between reviewers in our model, as captured by the game $\mathcal{G}_c$. The individualistic nature of the $\hat{S}^{i}_{p}$ estimation is what drives the need for the Nash equilibrium analysis. Reviewers must not only consider their perception of the paper's quality but also anticipate the reactions of the other reviewers to maximize their payoff.

\begin{Definition}
\label{def:qualityscore}
We define $\hat{S}^{i}_{p}$ as the estimated quality score of the paper $p$ by reviewer $i$, and $b^{i}_{p}$ as the bias of reviewer $i$. The bias represents the reviewer's personal judgment, which may deviate from the true quality score of the paper.
\end{Definition}

\begin{Remark}
\label{rmk:reviewerbias}
The bias term $b^{i}_{p}$ is a reflection of the inherent subjectivity in the review process. It may depend on factors such as the reviewer's expertise, personal preferences, or even their mood at the time of review. In real-world scenarios, this bias term can never be fully eliminated. However, by making the reward system more efficient, we can motivate reviewers to minimize the impact of their bias, which in turn improves the overall quality of the review process.
\end{Remark}

It is generally accepted that there are certain thresholds in the quality score that determine whether a paper should be accepted, requires revision, or should be rejected. We denote these thresholds as $Th_a$ and $Th_r$, respectively.

\begin{Definition}
\label{def:thresholds}
Let $Th_a$ and $Th_r$ be the quality score thresholds for acceptance and revision, respectively. A paper is deemed suitable for acceptance if its quality score is greater than $Th_a$; it requires revision if its quality score is between $Th_a$ and $Th_r$; and it is deemed suitable for rejection if its quality score is less than $Th_r$.
\end{Definition}

The quality score thresholds provide a quantitative benchmark against which the merit of a paper can be gauged. The thresholds effectively segregate papers into distinct categories, allowing for a more streamlined review process. These thresholds, though arbitrary, serve to uphold the standards of the journal.

\begin{Assumption}
\label{asmp:thresholds}
We assume that the thresholds $Th_a$ and $Th_r$ are such that $Th_a > Th_r$. This assumption is made to ensure that papers of higher quality are more likely to be accepted, while papers of lower quality are more likely to be rejected.
\end{Assumption}

\begin{Assumption}
\label{asmp:sharedthresholds}
We assume that the thresholds $Th_a$ and $Th_r$ are common knowledge among reviewers.
\end{Assumption}

\begin{Remark}
\label{rmk:decisionbasedonbiasedscore}
Given Assumption \ref{asmp:sharedthresholds}, each reviewer makes their decision based on $\hat{S}^{i}_{p}$, their own biased estimate of the paper's quality. This is their best strategy to maximize their expected voucher value, as they are trying to estimate the other reviewer's decision.
\end{Remark}

\begin{Assumption}
\label{asmp:rewardmaximization}
Although reviewers' decisions are influenced by several factors, in this paper, we assume that each reviewer aims to maximize their voucher value.
\end{Assumption}

In the ensuing examination, we delve deeper into the implications of the postulated Assumption \ref{asmp:rewardmaximization}, which asserts that a reviewer's primary objective is to maximize their voucher value. This assumption is made to provide a framework for analyzing the effect of the voucher system on the decision-making process of reviewers. It is pivotal to underscore that this assumption is a simplification of reality, as it abstracts away from other considerations that might influence a reviewer's decision.

\begin{Remark}
\label{rmk:assumptioncontext}
The assumption elucidated in Assumption \ref{asmp:rewardmaximization}, that a reviewer's decision-making is predominantly driven by the intent to augment their voucher value, serves primarily as a tool of simplification for the complexity inherent in the system under scrutiny. It is essential to interpret this assumption within the context of the theoretical framework we adopt in this paper, and not as a blanket statement on the motives of every reviewer in the open-access journal landscape.
\end{Remark}

We take a moment to emphasize the role of our assumption, as specified in Assumption \ref{asmp:rewardmaximization} and Remark \ref{rmk:assumptioncontext}, within the broader scope of our study. Our focus is on unveiling the strategic implications of the reward system, a task for which the assumption of reward maximization provides a tractable analytical lens. This assumption is not intended to be a definitive or comprehensive characterization of reviewers' motivations, which, in reality, are multifaceted and may encompass a commitment to scholarly rigor, the advancement of their field, and upholding the integrity of the scientific method.

\begin{Remark}
\label{rmk:assumptionlimits}
It is incumbent upon us, in our quest for mathematical rigor, to acknowledge the limits of the assumptions we make. In Assumption \ref{asmp:rewardmaximization}, we posit a certain degree of economic rationality in reviewers' behaviors. This does not, however, negate the potential for other non-economic factors to influence their decisions. Indeed, many reviewers are guided by a sense of duty to their academic community, and their decisions reflect a careful evaluation of the manuscript's quality and contribution to knowledge. Our assumption, therefore, is an abstraction that aids in the exploration of certain strategic dynamics within the reviewer reward system, rather than a comprehensive model of reviewer behavior.
\end{Remark}

The underpinning assumption, Assumption \ref{asmp:rewardmaximization}, can be perceived as a behavioral model, an economic perspective on the decision-making process of reviewers. The assumption postulates that each reviewer aims to maximize their voucher value, i.e., reviewers are essentially trying to maximize the expected present value of the voucher they receive. It is worth pointing out that this assumption does not necessarily signify that reviewers prioritize their self-interest over the integrity of the review process. Instead, it provides an economic rationale for the reviewer's decision-making process, which complements the existing understanding that the reviewer's decision is primarily driven by their assessment of the paper's quality.

In the light of this assumption, the decision-making process of a reviewer can be modeled as a form of strategic interaction where each reviewer makes their decision considering the potential decisions of the other reviewer, the quality of the paper, and the voucher system's time-dependent nature. The reviewer's decision-making process thus becomes an intricate dance between maintaining the integrity of the review process, assessing the quality of the paper, and maximizing the expected voucher value.

In the context of our study, the time-dependent nature of the voucher system introduces a dynamic that further complicates the reviewer's decision-making process. It is important to note that, under Assumption \ref{asmp:vouchertime}, the "Revision" decision effectively discounts the value of the voucher due to the time delay in its receipt, even though the face value of the voucher is the same for all three decisions. Hence, when a reviewer is trying to maximize their voucher value, they might be incentivized to choose either "Accept" or "Reject" over "Revision".

\section{Current Reviewer Reward System: A Game-Theoretic Analysis}
\label{sec:currentsystem}

\subsection{Model Formulation}
\label{subsec:modelformulation}

We begin the analysis by modeling the current reviewer reward system as a game. Recall from Section \ref{sec:preliminaries} that a game is defined by a triplet $(\mathcal{P}, \mathcal{S}, \mathcal{U})$, where $\mathcal{P}$ is the set of players, $\mathcal{S}$ is the strategy space, and $\mathcal{U}$ is the set of utility functions.

\begin{Definition}
\label{def:currentreviewgame}
The reviewer reward system game under the current system, denoted $\mathcal{G}_c$, is defined as:
\begin{itemize}
\item $\mathcal{P} = {1,2}$, the set of two reviewers.
\item $\mathcal{S} = \mathcal{D} \times \mathcal{D}$, where $\mathcal{D} = \{\text{Accept}, \text{Revision}, \text{Reject}\}$ is the set of decisions a reviewer can make (refer to Definition \ref{def:reviewdecisions}).
\item $\mathcal{U} = (U_1, U_2)$, where $U_i: \mathcal{S} \rightarrow \mathbb{R}$ is the utility function of reviewer $i$.
\end{itemize}
\end{Definition}

In this game, each reviewer independently chooses a decision $d \in \mathcal{D}$ based on their evaluation of the paper's quality. The utility of a reviewer is determined by the voucher value they receive, which depends on the decision they make and the time at which they receive the voucher (refer to Assumption \ref{asmp:vouchertime} and Remark \ref{rmk:discountedvalue}).

An example of the payoff matrix of this game, denoted $\mathcal{M}_c$, is as follows:

\begin{equation}
\label{eq:currentpayoffmatrix}
\mathcal{M}_c =
\begin{array}{cc|ccc}
        & &   & \textbf{Reviewer 1} &  \\
        & & \text{Accept} & \text{Revision} & \text{Reject} \\
        \hline
        & \text{Accept} & 100, 100 & 100-d_t, 100-d_t & 100-d_t, 100-d_t \\
\textbf{Reviewer 2} & \text{Revision} & 100-d_t, 100-d_t & 100-d_t, 100-d_t & 100-d_t, 100-d_t \\
        & \text{Reject} & 100-d_t, 100-d_t & 100-d_t, 100-d_t & 100, 100
\end{array}
\end{equation}
where the first number in each cell is the utility of reviewer 1, the second number is the utility of reviewer 2, and $d_t$ represents the time discount due to the delay in voucher receipt in the case of a "Revision" decision.

For instance, to calculate $d_t$, we consider the daily interest rate $r_d$ and the number of days of revision $n_d$. Specifically, $d_t$ can be computed as $d_t = (1 + r_d)^{n_d}$. This formulation captures the impact of delayed voucher receipt on the utilities of both reviewers in the "Revision" decision scenario.

The reviewers aim to maximize their utility, which, in this game, is equivalent to receiving the highest possible voucher value in the shortest possible time. They do this by estimating the other reviewer's decision based on their own assessment of the paper's quality, and choosing a decision that they believe is most likely to be in agreement with the other reviewer's decision. This is formalized in the following proposition:

\begin{Proposition}
\label{prop:strategy}
In the game $\mathcal{G}_c$, the dominant strategy for reviewer $i$ is to choose a decision $d \in \mathcal{D}$ such that $d = \arg\max_{d'} U_i(d', \hat{d}^{-i})$, where $\hat{d}^{-i}$ is reviewer $i$'s estimation of the other reviewer's decision.
\end{Proposition}

\begin{proof}
Given the payoff matrix $\mathcal{M}_c$ (Equation \ref{eq:currentpayoffmatrix}), the utility for reviewer $i$ when they choose decision $d$ and the other reviewer chooses $\hat{d}^{-i}$ is $U_i(d, \hat{d}^{-i})$. By choosing $d = \arg\max_{d'} U_i(d', \hat{d}^{-i})$, reviewer $i$ is maximizing their utility, given their estimation of the other reviewer's decision. This proves that choosing such a $d$ is a dominant strategy for reviewer $i$.
\end{proof}

\subsection{Nash Equilibrium Analysis}
\label{subsec:nashanalysis}

We now turn our attention to the analysis of Nash equilibriums for the game $\mathcal{G}_c$ defined in Definition \ref{def:currentreviewgame}. Nash equilibriums, as defined in Definition \ref{def:nash}, are strategy profiles from which no player has an incentive to unilaterally deviate, given the strategies of the other players.

\begin{Assumption}
\label{asmp:strategy}
Each reviewer $i$'s strategy is to choose the decision $d \in \mathcal{D}$ that maximizes their expected utility, given their estimated quality score $\hat{S}^{i}_{p}$ and the thresholds $Th_a$ and $Th_r$ from Assumption \ref{asmp:thresholds}.
\end{Assumption}

Let's denote the set of strategies that maximize the expected utility of reviewer $i$ given their estimated quality score $\hat{S}^{i}_{p}$ as $S^i(\hat{S}^{i}_{p})$. Reviewer $i$'s optimal strategy $s^*_i$ is then a member of this set.

\begin{Lemma}
\label{lem:strategynash}
Given the thresholds $Th_a$ and $Th_r$, a strategy profile $(s^*_1, s^*_2)$ is a Nash equilibrium of the game $\mathcal{G}_c$ if and only if $s^*_i \in S^i(\hat{S}^{i}_{p})$ for $i \in \{1,2\}$.
\end{Lemma}

Proof of this lemma follows directly from Assumption \ref{asmp:strategy} and the definition of a Nash equilibrium (Definition \ref{def:nash}). In particular, if $s^*_i \in S^i(\hat{S}^{i}_{p})$, then by definition, there does not exist a strategy $s_i \in S_i$ such that $U_i(s_i, s^*_{-i}) > U_i(s^*_i, s^*_{-i})$, which means that $(s^*_1, s^*_2)$ satisfies the condition for a Nash equilibrium.

\begin{Theorem}
\label{thm:ne-currentsystem}
For all $S_p \in \mathbb{R}$, there exists a Nash equilibrium $(s^*_1, s^*_2)$ only in (Accept, Accept) and (Reject, Reject) in the game $\mathcal{G}_c$.
\end{Theorem}

\begin{proof}
The proof of Theorem \ref{thm:ne-currentsystem} can be completed through an exhaustive analysis of the strategies available to each player under the conditions given by the assumptions and definitions. We aim to show that, for all $S_p \in \mathbb{R}$, the only Nash equilibrium strategies are (Accept, Accept) and (Reject, Reject).

Recall the strategy sets $S^i(\hat{S}^{i}_{p})$ and the optimal strategies $s^*_i \in S^i(\hat{S}^{i}_{p})$ as defined in Lemma \ref{lem:strategynash}. These strategy sets are determined by the thresholds $Th_a$ and $Th_r$ as well as the estimated quality scores $\hat{S}^{i}_{p}$ of the paper $p$ by reviewer $i$.

Assume that the thresholds $Th_a$ and $Th_r$ are fixed. Then, the estimated quality score $\hat{S}^{i}_{p}$ of the paper $p$ by reviewer $i$ uniquely determines $s^*i$. This is because $\hat{S}^{i}_{p}$ is an increasing function of $S_p + b^{i}_{p}$, and the reviewers will choose the decision that maximizes their expected utility.

Consider the three possible decisions for each player: Accept, Revision, and Reject. We will discuss each possibility in turn.

\begin{itemize}
\item If $\hat{S}^{i}_{p} > Th_a$, the reviewers will choose to accept the paper. In this case, both reviewers are in agreement, and the Nash equilibrium is (Accept, Accept).
\item If $Th_a > \hat{S}^{i}_{p} > Th_r$, the reviewers may choose to revise the paper. However, considering the payoff matrix $\mathcal{M}_c$ (Equation \ref{eq:currentpayoffmatrix}), choosing to revise does not maximize the reviewers' utility, as they face a time delay in receiving their vouchers, represented by $d_t$ in the payoff matrix. Instead, the reviewers will be motivated to unilaterally deviate from this decision to either accept or reject the paper. Therefore, (Revision, Revision) cannot be a Nash equilibrium.
\item If $Th_r > \hat{S}^{i}_{p}$, the reviewers will choose to reject the paper. Both reviewers are again in agreement, and the Nash equilibrium is (Reject, Reject).
\end{itemize}

In conclusion, for any inherent quality score $S_p \in \mathbb{R}$, the only Nash equilibrium strategies in the game $\mathcal{G}_c$ are (Accept, Accept) and (Reject, Reject). 
\end{proof}

\begin{Example}
\label{ex:ne-currentsystem}
Consider a scenario where $S_p = 0.7$, $b^{1}_{p} = -0.5$, and $b^{2}_{p} = 0.1$. Then, $\hat{S}^{1}_{p} = 0.65$ and $\hat{S}^{2}_{p} = 0.80$. Assuming $Th_a = 0.75$ and $Th_r = 0.50$, reviewer 1 would likely choose to revise the paper, while reviewer 2 would likely choose to accept it. However, to maximize their utility, they would try to align their decisions. Depending on their risk tolerance and other factors, they might both choose to revise or accept. Both of these strategy profiles, (Revision, Revision) and (Accept, Accept), could be Nash equilibriums of this game. Note that the choice between these equilibriums could significantly impact the outcome for the paper.
\end{Example}

\begin{Remark}
\label{rmk:ne-currentsystem}
In the current reviewer reward system, the Nash equilibrium could incentivize reviewers to align their decisions, potentially leading to less accurate evaluations of papers. This could be particularly problematic if the inherent quality score of a paper is close to the threshold for acceptance or rejection, as in the case of Example \ref{ex:ne-currentsystem}. This highlights the potential need for a more effective reviewer reward system, which we will explore in the next section.
\end{Remark}

Thus, our analysis shows that under the current reviewer reward system, there exists a Nash equilibrium for each paper. However, the alignment of reviewer decisions in these equilibriums could potentially compromise the quality of the review process.

\section{Proposed Reviewer Reward System: A Game-Theoretic Analysis}
\label{sec:proposedsystem}

\subsection{Model Formulation}
\label{subsec:proposedmodelformulation}

Having analyzed the current system, we now turn our attention to the proposed reviewer reward system, which we model as a strategic game.

\begin{Definition}
\label{def:proposedreviewgame}
The reviewer reward system game under the proposed system, denoted as $\mathcal{G}_p$, is defined as follows:
\begin{itemize}
\item $\mathcal{P} = {1,2}$, the set of two reviewers.
\item $\mathcal{S} = \mathcal{D} \times \mathcal{D}$, where $\mathcal{D} = {\text{Accept}, \text{Revision}, \text{Reject}}$ is the set of decisions a reviewer can make.
\item $\mathcal{U} = (U_1, U_2)$, where $U_i: \mathcal{S} \rightarrow \mathbb{R}$ is the utility function of reviewer $i$.
\end{itemize}
\end{Definition}

Similar to the current system, each reviewer independently chooses a decision $d \in \mathcal{D}$ based on their evaluation of the paper's quality. The main difference in the proposed system lies in the utility functions: the voucher value, which is the reviewers' utility, varies not only with their decision and the timing of voucher receipt but also with the agreement between the reviewers' decisions.

An example of the payoff matrix for this game, denoted $\mathcal{M}_p$, is given by:

\begin{equation}
\label{eq:proposedpayoffmatrix}
\mathcal{M}_p =
\begin{array}{cc|ccc}
        & &   & \textbf{Reviewer 1} &  \\
        & & \text{Accept} & \text{Revision} & \text{Reject} \\
        \hline
        & \text{Accept} & 200, 200 & 100, 100 & 50, 50 \\
\textbf{Reviewer 2}  & \text{Revision} & 100, 100 & 200, 200 & 100, 100 \\
        & \text{Reject} & 50, 50 & 100, 100 & 200, 200
\end{array}
\end{equation}
where the first number in each cell is the utility of reviewer 1 and the second number is the utility of reviewer 2.

Given the structure of the payoff matrix $\mathcal{M}_p$, the reviewers are incentivized to make their decisions not only based on their own assessments but also with an eye towards agreement with the other reviewer. This leads to the following proposition:

\begin{Proposition}
\label{prop:proposedstrategy}
In the game $\mathcal{G}_p$, the optimal strategy for reviewer $i$ is to choose a decision $d \in \mathcal{D}$ such that $d = \arg\max_{d'} U_i(d', \hat{d}^{-i})$, where $\hat{d}^{-i}$ is reviewer $i$'s estimation of the other reviewer's decision.
\end{Proposition}

\begin{proof}
The proof of Proposition \ref{prop:proposedstrategy} is similar to that of Proposition \ref{prop:strategy} in Section \ref{subsec:modelformulation}. Given the payoff matrix $\mathcal{M}_p$ (Equation \ref{eq:proposedpayoffmatrix}), the utility for reviewer $i$ when they choose decision $d$ and the other reviewer is the utility of reviewer 2.
\end{proof}

The underlying principles in evaluating the quality of the paper remain the same as in the previous game, as detailed in Assumption \ref{asmp:thresholds} and Definition \ref{def:qualityscore}. However, given the modified payoff matrix $\mathcal{M}_p$, the strategy for each reviewer changes.

\begin{Proposition}
\label{prop:strategy_proposed}
In the game $\mathcal{G}_p$, the strategy for reviewer $i$ is to choose a decision $d \in \mathcal{D}$ such that $d = \arg\max_{d'} U_i(d', \hat{d}^{-i})$, where $\hat{d}^{-i}$ is reviewer $i$'s estimation of the other reviewer's decision.
\end{Proposition}

\begin{proof}
Given the payoff matrix $\mathcal{M}_p$ (Equation \ref{eq:proposedpayoffmatrix}), the utility for reviewer $i$ when they choose decision $d$ and the other reviewer chooses $\hat{d}^{-i}$ is $U_i(d, \hat{d}^{-i})$. By choosing $d = \arg\max_{d'} U_i(d', \hat{d}^{-i})$, reviewer $i$ is maximizing their utility, given their estimation of the other reviewer's decision. This proves that choosing such a $d$ is a rational strategy for reviewer $i$ in the proposed system.
\end{proof}

\begin{Remark}
\label{rmk:decisionchange}
We observe that the change in the payoff matrix affects the decision-making process of the reviewers. Reviewers are now incentivized to more accurately assess the paper and coordinate their decision with the other reviewer, as the highest reward is received when both make the same decision.
\end{Remark}

\begin{Definition}
\label{def:reward}
We define the reward function $R: \mathcal{S} \rightarrow \mathbb{R}$ for the proposed system as $R(d_1, d_2) = \mathcal{M}_p[d_1, d_2]$, where $d_1, d_2 \in \mathcal{D}$ are the decisions of the reviewers.
\end{Definition}

In the proposed system, the reward function is symmetric, i.e., $R(d_1, d_2) = R(d_2, d_1)$ for all $d_1, d_2 \in \mathcal{D}$. This symmetry in rewards encourages cooperation between reviewers, driving them to reach a consensus on the quality of the paper. The proposed system thus fosters a more collaborative and quality-focused review process.

\begin{Assumption}
\label{asmp:belief}
We assume that reviewer $i$ believes that the other reviewer, $i'$, makes their decision based on their estimated quality score, $\hat{S}^{i'}_{p}$.
\end{Assumption}

This assumption adds another layer of complexity to the reviewers' decision-making process. Under this assumption, reviewer $i$ must not only consider their own evaluation of the paper, but also estimate the decision of the other reviewer, $i'$, based on their own quality assessment.

Given that reviewer $i$ only has the paper as a source of information, the most reasonable estimation they can make about $\hat{S}^{i'}{p}$ is that it equals their own estimated quality score, $\hat{S}^{i}_{p}$. This leads to the assumption that $\hat{S}^{i'}_{p} \simeq \hat{S}^{i}_{p}$. In other words, in the absence of any other information, reviewer $i$ assumes that the other reviewer's evaluation of the paper aligns with their own.

This assumption holds some basis in reality. When a reviewer is evaluating a paper, they may naturally assume that another reviewer, with a similar level of expertise and understanding, would evaluate the paper similarly. Indeed, this may not always be the case due to differing biases and perspectives, but in the absence of other information, this serves as a reasonable assumption.

Under this assumption, each reviewer tries to estimate the other's decision based on their own assessment of the paper. Consequently, this encourages reviewers to more closely scrutinize the paper, as their reward is now dependent on not only their own decision but also on accurately predicting the other reviewer's decision.

Therefore, in the proposed system, we see that the incentives are designed in such a way as to encourage both accurate evaluation of the paper and coordination between reviewers. This could potentially lead to an increase in the quality of the reviews, and ultimately, of the papers accepted by the journal.

\subsection{Nash Equilibrium Analysis}
\label{subsec:nashanalysis_proposed}

We start by considering the Nash Equilibrium of the proposed game $\mathcal{G}_p$.

\begin{Definition}
\label{def:nash_proposed}
A strategy profile $(s^*_1, s^*_2)$ is a Nash equilibrium of the proposed game $\mathcal{G}_p$ if and only if for every player $i \in \{1,2\}$, we have $U_i(s^*_i, s^*_{-i}) \geq U_i(s_i, s^*_{-i})$ for all $s_i \in S_i$.
\end{Definition}

This definition follows directly from the standard definition of Nash equilibrium. In particular, a strategy profile is a Nash equilibrium if no player can unilaterally improve their utility by deviating from their current strategy.

\begin{Proposition}
\label{prop:nash_proposed}
In the proposed game $\mathcal{G}_p$, a strategy profile $(s^*_1, s^*_2)$ is a Nash equilibrium if and only if $s^*_i \in S^i(\hat{S}^{i}_{p})$ for $i \in \{1,2\}$, where $S^i(\hat{S}^{i}_{p})$ is the set of optimal strategies for player $i$ given their estimated quality score $\hat{S}^{i}_{p}$.
\end{Proposition}

\begin{proof}
The proof of Proposition \ref{prop:nash_proposed} is similar to that of Lemma \ref{lem:strategynash} for the current system. Specifically, if $s^i \in S^i(\hat{S}^{i}_{p})$, then by definition of $S^i(\hat{S}^{i}_{p})$, there does not exist a strategy $s_i \in S_i$ such that $U_i(s_i, s^*_{-i}) > U_i(s^*_i, s^*_{-i})$, which means that $(s^*_1, s^*_2)$ satisfies the condition for a Nash equilibrium.
\end{proof}

\begin{Theorem}
\label{thm:ne_proposedsystem}
For all $S_p \in \mathbb{R}$, there exists a Nash equilibrium $(s^*_1, s^*_2)$ in the game $\mathcal{G}_p$.
\end{Theorem}

\begin{proof}
Given $S_p$, the estimated quality score by each reviewer $i$ is $\hat{S}^{i}_{p} = S_p + b^{i}_{p}$. From Proposition \ref{prop:strategy_proposed}, we know that each reviewer $i$ chooses a decision $d \in \mathcal{D}$ that maximizes their expected utility, which results in a strategy $s^i \in S^i(\hat{S}^{i}_{p})$. From Proposition \ref{prop:nash_proposed}, we know that if $s^i \in S^i(\hat{S}^{i}_{p})$ for $i \in \{1,2\}$, then $(s^*_1, s^*_2)$ is a Nash equilibrium of the game $\mathcal{G}_p$. Therefore, we can conclude that for all $S_p \in \mathbb{R}$, there exists a Nash equilibrium $(s^*_1, s^*_2)$ in the game $\mathcal{G}_p$.
\end{proof}

\begin{Remark}
\label{rem:strategyspace_proposed}
In the proposed game $\mathcal{G}_p$, the strategy space $S_i$ of a player $i$ is the set of all possible decisions, $S_i = \{Accept, Revision, Reject\}$. However, the effective strategy space $S^i(\hat{S}^{i}_{p})$ can be much smaller depending on the estimated quality score $\hat{S}^{i}_{p}$. In particular, if $\hat{S}^{i}_{p}$ is close to either end of the spectrum, the effective strategy space can be reduced to a singleton set, making the player's decision trivial.
\end{Remark}

\begin{Example}
\label{ex:nash_proposed}
Consider a paper with an inherent quality score $S_p = 0.7$. Let's assume that reviewer 1 and reviewer 2 have biases $b^1_p = 0.1$ and $b^2_p = -0.1$, respectively. Therefore, the estimated quality scores by the two reviewers are $\hat{S}^1_p = 0.8$ and $\hat{S}^2_p = 0.6$. If the thresholds are $Th_a = 0.75$ and $Th_r = 0.5$, then the optimal strategy for reviewer 1 is to accept the paper, while the optimal strategy for reviewer 2 is to suggest a revision. In this case, the Nash equilibrium of the game is $(Accept, Revision)$, which results in a reward of 100 for each reviewer according to the proposed reward system.
\end{Example}

\begin{Definition}
\label{def:efficiency_proposed}
A strategy profile $(s^*_1, s^*_2)$ is Pareto efficient in the proposed game $\mathcal{G}_p$ if and only if there does not exist another strategy profile $(s_1, s_2)$ such that $U_i(s_i, s_{-i}) \geq U_i(s^*_i, s^*_{-i})$ for all $i \in \{1,2\}$, and $U_i(s_i, s_{-i}) > U_i(s^*_i, s^*_{-i})$ for at least one $i$.
\end{Definition}

\begin{Proposition}
\label{prop:efficiency_proposed}
In the proposed game $\mathcal{G}_p$, a strategy profile $(s^*_1, s^*_2)$ is Pareto efficient if and only if $s^*_i = \arg\max_{s_i \in S_i} U_i(s_i, s^*_{-i})$ for $i \in \{1,2\}$.
\end{Proposition}

The analysis of the Nash equilibria in the proposed game $\mathcal{G}_p$ reveals that the game encourages reviewers to make accurate estimations of the quality of a paper and to choose their decisions accordingly.

\section{Simulations and Results}
\label{sec:simulationsandresults}

In this section, we explore the characteristics and attributes of the existing reviewer reward system by conducting simulations. The simulations are performed using a straightforward deep reinforcement learning approach, in which two agents are employed to represent the reviewers. The primary objective of these agents is to maximize their anticipated utility within the game $\mathcal{G}_c$, as defined in Definition \ref{def:currentreviewgame}. It is important to note that the focus of this paper does not lie in the intricate details of the deep reinforcement model architecture; hence, we employ a simplified model for our simulations.

\subsection{Simulation Settings of Current Reviewer Reward System}
\label{subsec:simulationsettings_current}

\begin{figure}[h!]
\centering
\includegraphics[width=\linewidth]{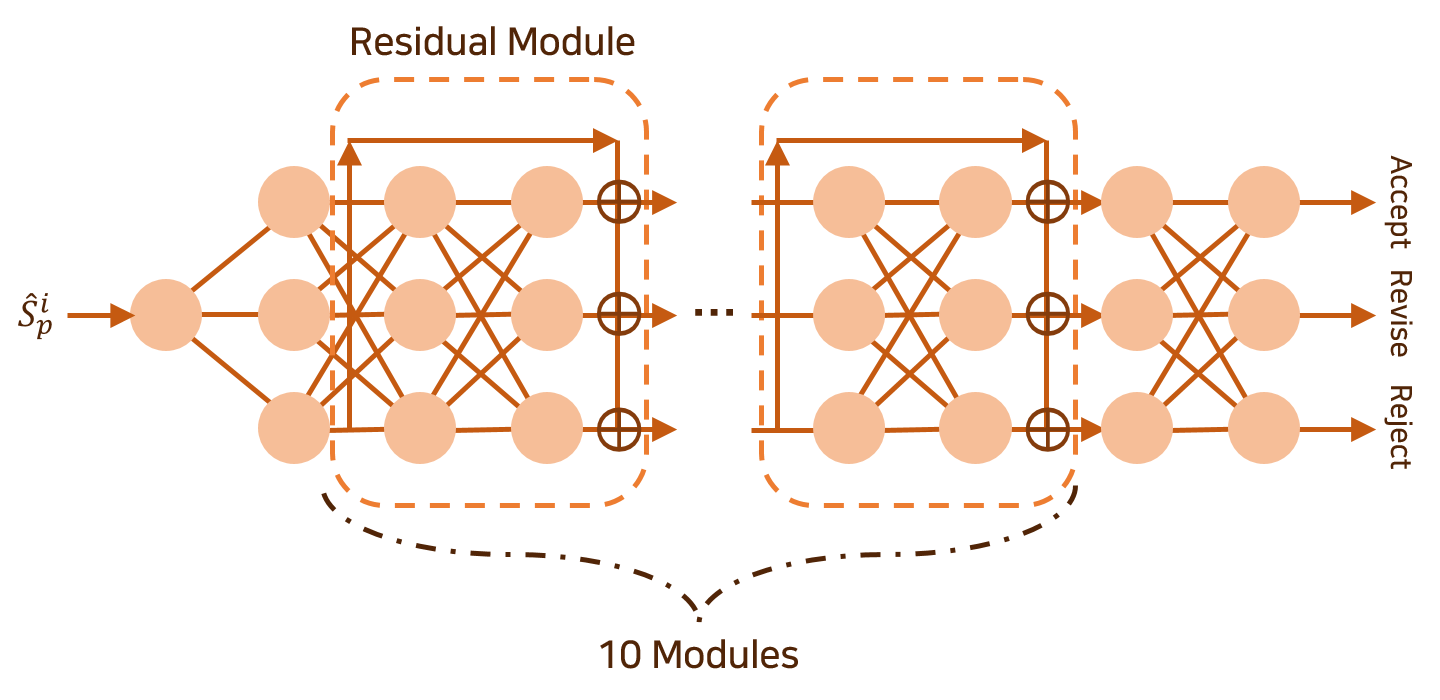}
\caption{Deep reinforcement learning structure of the agents.}
\label{fig:agentnn}
\end{figure}

The two agents are modelled as neural networks with a single input, three outputs, and a hidden layer of size 256. The input to each agent is the estimated quality score $\hat{S}^{i}_{p}$ of a paper, while the output represents the probability distribution over the set of possible decisions $\mathcal{D} = \{\text{Accept}, \text{Revision}, \text{Reject}\}$. The structure of the agents' neural networks is depicted in Figure \ref{fig:agentnn}.

In addition, each neural network employs a series of ten residual modules \cite{he2016deep} to facilitate the training process. These modules are designed to learn identity functions that allow the input signals to bypass some layers and reach deeper parts of the network directly, effectively addressing the issue of vanishing gradients commonly observed in deep neural networks \cite{paul2015review}.

The agents' neural networks are trained using the Adam optimizer \cite{kingma2014adam} with a learning rate of $0.00001$, and the training process lasts for $100,000$ epochs. The reward at each epoch is calculated based on whether the decisions made by the agents are in agreement and whether these decisions are correct. Specifically, the agents receive a reward of $1.0$ if they both make the correct decision (either accepting or rejecting the paper), and a reward of $1.0 - d_t$ if they make different decisions or both decide to request a revision. The quantity $d_t$ is a time discount factor set to $0.2$, which represents the delay in voucher receipt for a "Revision" decision.

In order to ensure that the agents' decisions are not solely based on the maximization of immediate reward, an auxiliary loss term is included in the loss function. This term is calculated as the cross-entropy between the agents' decisions and the true decision, and serves to guide the agents towards making decisions that are consistent with the actual quality of the papers.

The quality score $S_p$ and the biases $b^{i}_{p}$ are randomly generated for each epoch. The quality score is drawn from a uniform distribution between $0$ and $1$, while the biases are drawn from a uniform distribution between $-0.05$ and $0.05$. The thresholds $Th_a$ and $Th_r$ for making "Accept" and "Revision" decisions are set to $0.66$ and $0.33$, respectively, in order to have the same probabilities of "Accept", "Revision", and "Reject" for a random paper, in accordance with Assumptions \ref{asmp:thresholds} and \ref{asmp:sharedthresholds}.

In order to observe the evolution of the agents' decision-making process during the training, we record the moving average of the rewards received by the agents, as well as the distribution of their decisions. The moving average reward is calculated over a sliding window of $100$ epochs, and the decision distribution is computed by counting the number of times each pair of decisions is made by the agents.

\subsection{Definition of Reward and Loss Functions}
\label{subsec:reward_loss_functions}

\subsubsection{Reward Function}
\label{subsec:reward_function}

The reinforcement learning framework for our simulation necessitates a reward function, which facilitates the process of decision-making. As the agents interact with the environment, they receive rewards that depend on the actions they take. The reward function we define here is constructed to encourage a pair of decisions that are both accurate and concordant.

We denote the reward at time $t$ as $R_t$. Let $d^i_t \in \mathcal{D}$ represent the decision made by agent $i$ at time $t$, where $\mathcal{D} = \{\text{Accept}, \text{Revision}, \text{Reject}\}$. The reward is defined as follows:

\begin{equation}
\label{eqn:reward_function}
R_t =
\begin{cases}
1.0 & \text{if } d^1_t = d^2_t \text{ and } (d^i_t = \text{Accept} \text{ or } d^i_t = \text{Reject}), \\
1.0 - d_t & \text{otherwise}.
\end{cases}
\end{equation}

The quantity $d_t$ is a time discount factor set to $0.2$, which represents the delay in voucher receipt for a "Revision" decision. This reward structure encourages the agents to agree with each other and to make correct decisions.

\subsubsection{Loss Function}
\label{subsec:loss_function}

Training our agents involves minimizing a loss function, which is composed of two parts: the expected reward and an auxiliary term. The expected reward, $-\mathbb{E}[R_t]$, encourages the agents to make decisions that maximize their reward. The auxiliary term, $L^{aux}_t$, is introduced to ensure that the agents' decisions align with the assessed quality of the papers. This auxiliary loss term is calculated as the cross-entropy between the agents' decisions and the true decision.

For agent $i$, the loss function at time $t$ is defined as follows:

\begin{equation}
\label{eqn:loss_function}
L^i_t = -\mathbb{E}[R_t] + \lambda L^{aux}_t,
\end{equation}
where $\lambda$ is a hyperparameter balancing the contribution of the auxiliary term. In our implementation, we set $\lambda = 0.05$.

The auxiliary loss for agent $i$ is calculated as:

\begin{equation}
\label{eqn:aux_loss}
L^{aux}_t = -\log p^i_{t}(d^i_t),
\end{equation}
where $p^i_{t}(d^i_t)$ is the probability assigned by agent $i$ to its decision $d^i_t$ at time $t$.

\begin{Remark}
We use the cross-entropy loss as our auxiliary term because it measures the dissimilarity between the predicted probability distribution and the target distribution. By minimizing this term, we encourage our agents to make decisions that align with the actual quality of the papers.
\end{Remark}

\subsection{Simulation Settings of Proposed Reviewer Reward System}
\label{subsec:simulationsettings_proposed}

The agents' architectures and the simulation framework remain the same for the proposed reviewer reward system as in the current system, as detailed in Subsection \ref{subsec:simulationsettings_current}. However, the crucial distinction lies in the reward function, which we redefine in this context to better incentivize the reviewers' decision-making process.

\begin{Definition}[Proposed Reward Function]
\label{def:proposed_reward_function}

Given that $d^i_t \in \mathcal{D}$ represents the decision made by agent $i$ at time $t$, where $\mathcal{D} = {\text{Accept}, \text{Revision}, \text{Reject}}$, we define the new reward at time $t$ as $R'_t$. This reward is characterized as follows:

\begin{equation}
\label{eqn:proposed_reward_function}
R'_t =
\begin{cases}
1.0 & \text{if } d^1_t = d^2_t, \\
0.5 & \text{if } (d^{i}_t, d^{i'}_t) = (\text{Accept, Revision}) \text{ or } (\text{Revision, Reject}), \\
0.2 & \text{otherwise}.
\end{cases}
\end{equation}
\end{Definition}

\begin{Remark}
\label{remark:proposed_reward}
In the proposed reward function, the agents get a reward of $1.0$ if their decisions agree and are correct, a reward of $0.5$ if their decisions are off by one category, and a reward of $0.2$ otherwise. This grading system is designed to encourage agreement among the agents, while at the same time accommodating minor discrepancies.
\end{Remark}

We employ this newly formulated reward function in our simulation, replacing the former reward function detailed in Subsection \ref{subsec:reward_function}. The loss function defined in Equation \ref{eqn:loss_function} remains unaltered for this experiment, as it provides a stable and consistent means of optimizing the agents' actions.

\begin{figure}[h!]
\centering
\includegraphics[width=\linewidth]{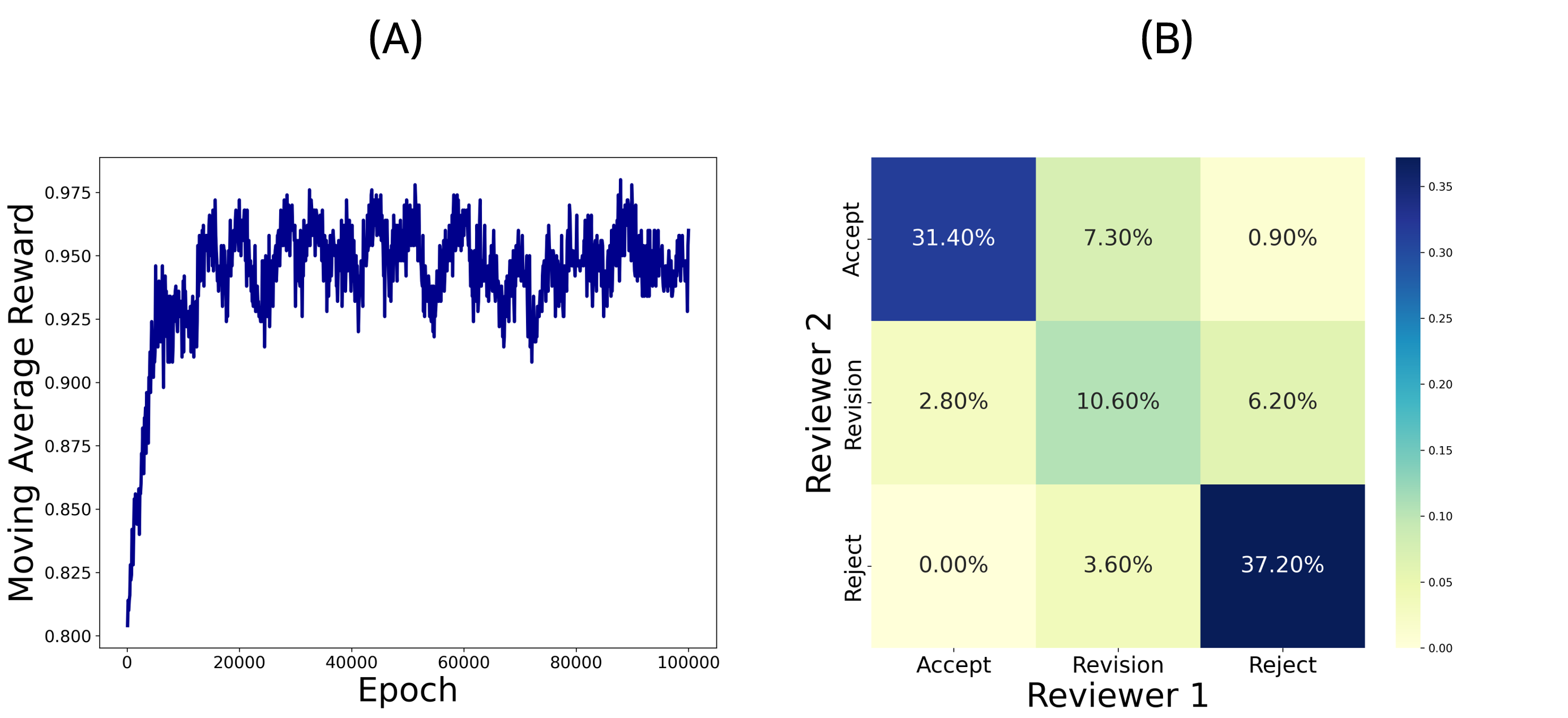}
\caption{Results of the current reviewer reward system. (A) Rewards over epoch. (B) Decision distributions of Reviewer 1 and 2.}
\label{fig:fig2}
\end{figure}

\subsection{Outcome Analysis and Interpretation}
\label{subsec:results}

\begin{figure}[h!]
\centering
\includegraphics[width=\linewidth]{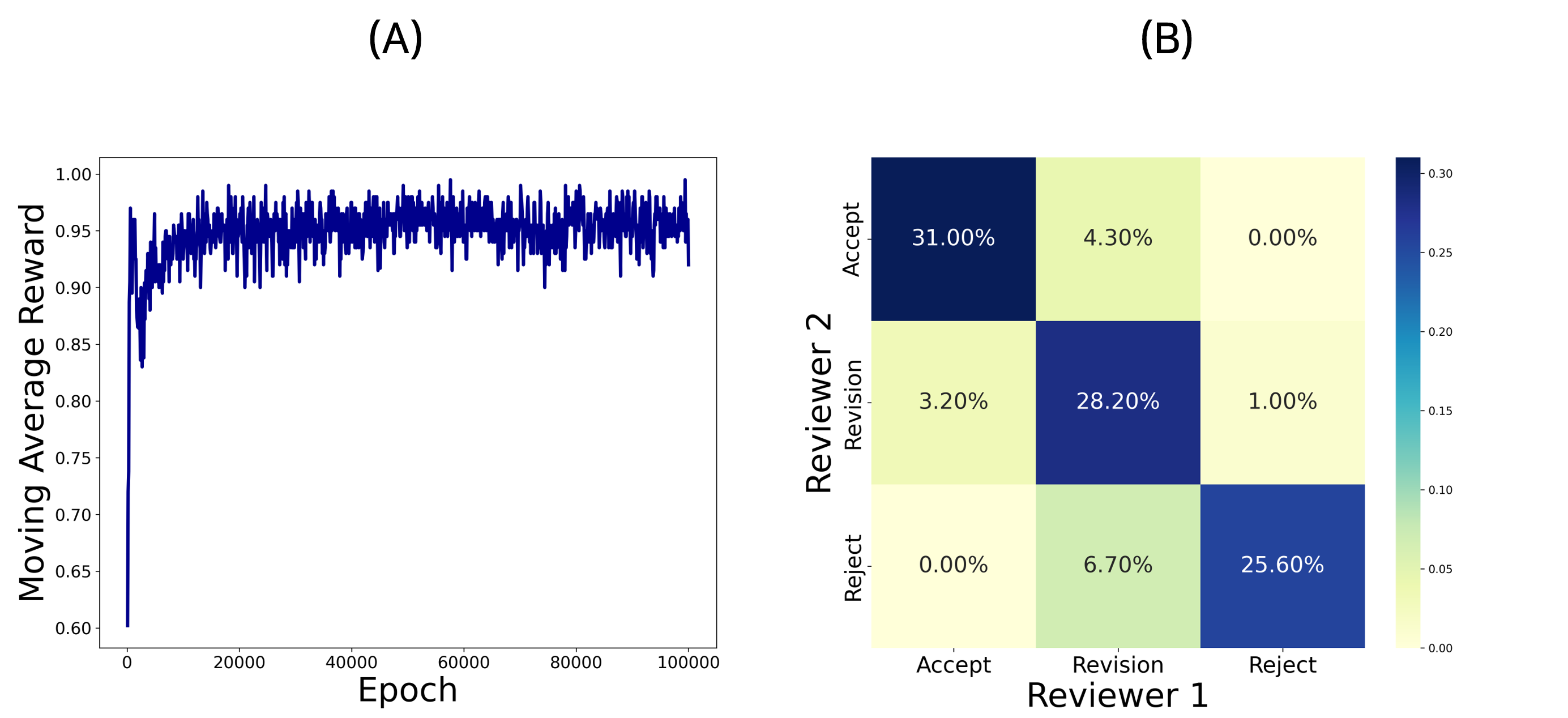}
\caption{Results of the proposed reviewer reward system. (A) Rewards over epoch. (B) Decision distributions of Reviewer 1 and 2.}
\label{fig:fig3}
\end{figure}

The results of the simulations, as showcased in Figures \ref{fig:fig2} and \ref{fig:fig3}, offer valuable insights into the performance of the current and proposed reviewer reward systems, respectively. In this subsection, we delve into these outcomes to distill the key findings and implications.

After a sequence of 20,000 epochs, the agents stabilize towards a Nash equilibrium, as discerned from the fact that the rewards settle around the value of $0.95$ and do not exhibit significant variations henceforth. The existence of Nash equilibria is predominantly evident in the pairings of (Accept, Accept) and (Reject, Reject), as illustrated in Figure \ref{fig:fig2}. In particular, the (Accept, Accept) combination is realized with a frequency of $31.40\%$, and the (Reject, Reject) pairing manifests itself with a frequency of $37.20\%$. Conversely, the (Revision, Revision) pairing is less common, with a likelihood of $10.60\%$. This pattern indicates that under the current reward system, reviewers have a propensity to adopt a binary decision-making approach, favoring either Accept or Reject verdicts.

In stark contrast, the proposed reward system demonstrates a more balanced distribution of decision pairings. As evidenced in Figure \ref{fig:fig3}, Nash equilibria can be found in all decision pairings, including (Accept, Accept), (Revision, Revision), and (Reject, Reject). In terms of percentages, these pairings are realized with probabilities of $31.00\%$, $28.20\%$, and $25.60\%$, respectively. Given the thresholds $Th_a$ and $Th_r$ set at $0.66$ and $0.33$ for "Accept" and "Revision" decisions, the ideal probabilities for the respective judgments ought to be around $33.33\%$. Consequently, the proposed reward system is more conducive to achieving this desirable state, demonstrating a more balanced approach to the review process compared to the current system.

We begin by observing the deviation of decision pairing frequencies from the idealized probabilities under both systems. For the current system, the deviation for (Accept, Accept) pairing, denoted as $\Delta_{AA}^c$, can be calculated as $|31.40\% - 33.33\%| = 1.93\%$, and similarly, $\Delta_{RR}^c = 3.87\%$, and $\Delta_{RevRev}^c = 22.73\%$. On the other hand, the proposed system exhibits the deviations as $\Delta_{AA}^p = 2.33\%$, $\Delta_{RR}^p = 7.73\%$, and $\Delta_{RevRev}^p = 5.13\%$. The significant reduction in deviation for the (Revision, Revision) pairing from $22.73\%$ to $5.13\%$ suggests a more balanced distribution of decision pairings under the proposed system.

A deeper analysis of these deviations reveals a noteworthy trend. Let $\Delta_{max}^c = \max\{\Delta_{AA}^c, \Delta_{RR}^c, \Delta_{RevRev}^c\}$ and $\Delta_{max}^p = \max\{\Delta_{AA}^p, \Delta_{RR}^p, \Delta_{RevRev}^p\}$ be the maximum deviations under the current and proposed systems, respectively. We observe that $\Delta_{max}^c > \Delta_{max}^p$, indicating that the proposed system brings the decision pairings closer to the idealized probabilities.

Another crucial observation pertains to the stability of the Nash equilibrium. By investigating the standard deviations of the rewards, denoted as $\sigma^c$ and $\sigma^p$ for the current and proposed systems respectively, we find that $\sigma^c > \sigma^p$. This suggests that the rewards are more stable under the proposed system, implying a more stable Nash equilibrium.

\section{Conclusion}
\label{sec:conclusion}

This study commenced with a comprehensive mathematical and game-theoretical examination of the reviewer reward systems prevalent in open-access journals. The existing system, construed as a strategic interaction between two reviewers, was discovered to engender a bias towards binary decisions. This bias was evidenced by the preponderance of Nash equilibria in (Accept, Accept) and (Reject, Reject) pairings, a trend that might imperil the thoroughness of peer reviews.

To counter this, we designed and examined an alternative reward system, intended to foster more balanced and comprehensive reviews. Through a combination of rigorous game-theoretical analysis and deep reinforcement learning simulations, we demonstrated the merits of our proposed system in fostering a more balanced and stable distribution of decisions. Our findings revealed that our proposed system nudges the distribution of decision pairings closer to the ideal probabilities, thereby promoting a more detailed review process.

However, it is crucial to acknowledge the inherent limitations of our model, which is predicated on Assumption \ref{asmp:rewardmaximization}. The presumption that reviewers primarily act to maximize their voucher value offers a simplified, yet insightful perspective, enabling us to isolate and scrutinize the influence of the reward system on decision-making. Yet, it overlooks a myriad of other factors that could potentially sway a reviewer's decision-making process. We concede that our model may not fully encapsulate the complexity and diversity of motivations inherent in the real-world peer review process. Consequently, our study should be viewed as an initial step rather than an endpoint. It paves the way for further research, encouraging subsequent studies to develop more sophisticated models of reviewer behavior that take into account the balance between voucher value maximization and other influencing factors. Such an endeavor could possibly furnish a more holistic understanding of the dynamics in the peer review process, enhancing the models we propose here and improving their practical relevance.

Our research offers a novel viewpoint on the topic of reviewer rewards in open-access journals, supplying a solid mathematical foundation to steer future research and policy deliberations. Although our findings hold promise, additional efforts should be dedicated to testing our proposed system in real-world contexts and assessing its practicality and effectiveness. Moreover, our methodology could be broadened to scrutinize other facets of the peer review process, thereby enriching the mathematical comprehension of this indispensable aspect of scholarly publishing.

\bibliographystyle{unsrt}
\bibliography{ref.bib}

\end{document}